\documentclass[sigconf,screen]{acmart}

\AtBeginDocument{%
  }

\usepackage{svg}

\usepackage{amssymb}

\usepackage{enumitem}
\usepackage{multirow}
\usepackage{fontawesome5}
\usepackage{diagbox}
\usepackage{colortbl}
\usepackage{makecell}
\usepackage{mdframed}
\usepackage{xspace}
\usepackage{threeparttable}
\usepackage{url}
\usepackage{caption}
\usepackage{booktabs}
\usepackage{subfig}
\usepackage{graphicx}
\usepackage{xcolor} 
\usepackage{pifont}
\usepackage{balance}
\usepackage{amsmath}
\usepackage{booktabs} 
\usepackage{array} 
\usepackage{CJKutf8}
\usepackage{adjustbox} 
\usepackage[most]{tcolorbox} 
\usepackage{tabularx}
\usepackage{array}
\usepackage{graphicx}
\usepackage{dashrule}
\usepackage{etoolbox}

\newcommand{\ie}{\textit{i}.\textit{e}.}

\newcommand{\model}{\texttt{MVGR-Net}}

\newcommand{\pink}[1]{{\textcolor{magenta}{\textbf{#1}}}}
\newcommand{\orange}[1]{{\textcolor{orange}{\textbf{#1}}}}
\copyrightyear{2026}
\acmYear{2026}
\setcopyright{cc}
\setcctype{by}
\acmConference[WWW '26] {Proceedings of the ACM Web Conference 2026}{April 13--17, 2026}{Dubai, United Arab Emirates.}
\acmBooktitle{Proceedings of the ACM Web Conference 2026 (WWW '26), April 13--17, 2026, Dubai, United Arab Emirates}
\acmISBN{979-8-4007-2307-0/2026/04}
\acmDOI{10.1145/3774904.3792789}

\begin{document}



\title[MVGR-Net]{Enhancing Ride-Hailing Forecasting at DiDi with Multi-View Geospatial Representation Learning from the Web}





\makeatletter
\renewcommand{\@fnsymbol}[1]{%
  \ensuremath{%
    \ifcase#1\or \dagger\or \ddagger\or \mathsection\or \mathparagraph\or 
    \|\or **\or \dagger\dagger\or \ddagger\ddagger\else\@ctrerr\fi
  }%
}
\makeatother

\author{Xixuan Hao}
\authornote{Equal Contribution.}
\affiliation{%
  \institution{The Hong Kong University of Science and Technology (Guangzhou)}
  \city{Guangzhou}
  \country{China}}
\email{xhao390@connect.hkust-gz.edu.cn}

\author{Guicheng Li}
\authornotemark[1]
\affiliation{%
  \institution{China University of Geoscience (Wuhan)}
  \city{Wuhan}
  \country{China}}
\email{liguicheng@cug.edu.cn}

\author{Daiqiang Wu}
\affiliation{%
  \institution{Didichuxing Co.\ Ltd}
  \city{Beijing}
  \country{China}
}
\email{wudaiqiang@didiglobal.com}

\author{Xusen Guo}
\affiliation{%
  \institution{The Hong Kong University of Science and Technology (Guangzhou)}
  \city{Guangzhou}
  \country{China}
}
\email{xguo796@connect.hkust-gz.edu.cn}

\author{Yumeng Zhu}
\affiliation{%
  \institution{Didichuxing Co.\ Ltd}
  \city{Beijing}
  \country{China}
}
\email{zhuyumeng@didiglobal.com}

\author{Zhichao Zou}
\affiliation{%
  \institution{Didichuxing Co.\ Ltd}
  \city{Beijing}
  \country{China}
}
\email{zouzhichao@didiglobal.com}

\author{Peng Zhen}
\affiliation{%
  \institution{Didichuxing Co.\ Ltd}
  \city{Beijing}
  \country{China}
}
\email{zhenpeng@didiglobal.com}

\author{Yao Yao}
\affiliation{%
  \institution{China University of Geoscience (Wuhan)}
  \city{Wuhan}
  \country{China}
}
\email{yaoy@cug.edu.cn}

\author{Yuxuan Liang}
\authornote{Corresponding author. Email: yuxliang@outlook.com}
\affiliation{%
  \institution{The Hong Kong University of Science and Technology (Guangzhou)}
  \city{Guangzhou}
  \country{China}
}
\email{yuxliang@outlook.com}

\renewcommand{\shortauthors}{Xixuan Hao et al.}

\definecolor{deeppink}{RGB}{255,20,147}
\begin{abstract}
The proliferation of ride-hailing services has fundamentally transformed urban mobility patterns, making accurate ride-hailing forecasting crucial for optimizing passenger experience and urban transportation efficiency. However, ride-hailing forecasting faces significant challenges due to geospatial heterogeneity and high susceptibility to external events. This paper proposes~\model~(\underline{M}ulti-\underline{V}iew \underline{G}eospatial \underline{R}epresentation Learning), a novel framework that addresses these challenges through a two-stage approach. In the pretraining stage, we learn comprehensive geospatial representations by integrating Points-of-Interest and temporal mobility patterns to capture regional characteristics from both semantic attribute and temporal mobility pattern views. The forecasting stage leverages these representations through a prompt-empowered framework that fine-tunes Large Language Models while incorporating external events. 
Extensive experiments on DiDi's real-world datasets demonstrate the state-of-the-art performance.
\end{abstract}





\maketitle

\vspace{-1em}
\begin{figure}[ht]
    \centering
\includegraphics[width=\linewidth]
    {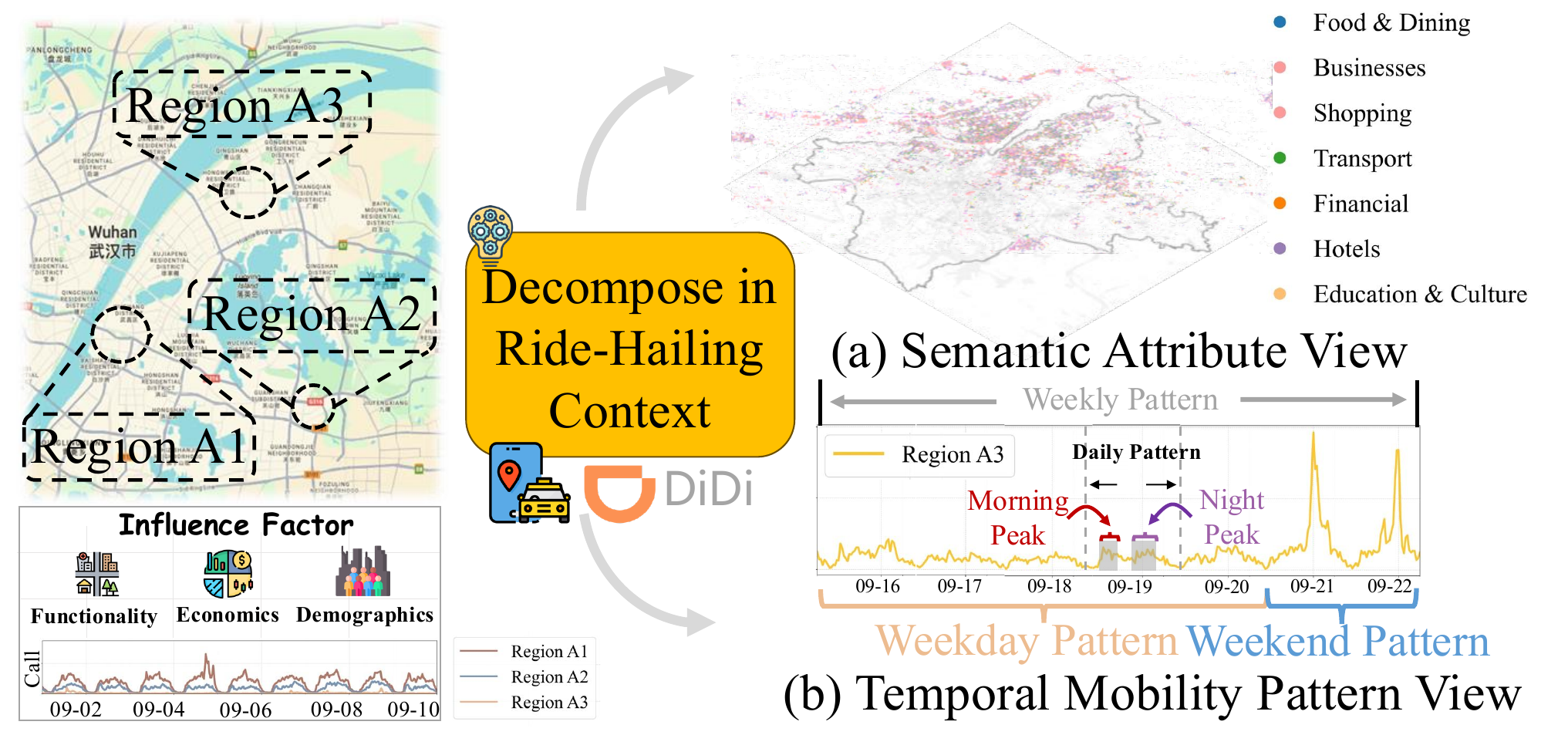}
    \caption{Geospatial Heterogeneity and its two principal view in the context of ride-hailing services.}
    \label{fig:intro}
    \vspace{-2em}
\end{figure}

\section{Introduction}
\label{text:intro}
With the rapid expansion of the World Wide Web, interconnectivity has induced a profound transformation in the way humans live. 
The subsequent emergence and rapid development of mobile Internet have given rise to a multitude of online services, among which ride-hailing platforms serve as prominent representatives.
As an exemplary web-data-driven service, the proliferation of ride-hailing has fundamentally reshaped urban mobility patterns. It has significantly enhanced transportation accessibility and flexibility, while simultaneously catalyzing innovations in intelligent transportation systems (ITS) and urban governance.
Ride-hailing forecasting~\cite{wgnn,stmgcn,ccrnn,2023Exploring} endeavors to predict temporal dynamics of key supply-demand indicators, such as Calls and Total Supply Hours (TSH), by leveraging historical transactional data in conjunction with exogenous variables including seasonal holidays and meteorological conditions.
Accurate ride-hailing forecasting enables optimization of passenger experience, enhancement of platform operational efficiency, mitigation of traffic congestion, and advancement of urban intelligent transportation ecosystems.

However, ride-hailing indicators, due to their close relationship
with urban mobility patterns, exhibit unique inherent characteristics
such as significant geospatial heterogeneity, and high susceptibility to external events~\cite{wgnn,refine,stresnet}. These properties render the task of achieving accurate ride-hailing forecasting a formidable challenge.
Existing studies~\cite{jin2020urban,lin2023deep,li2024optimization,zhang2021mlrnn}  commonly formulate ride-hailing
forecasting as a spatio-temporal prediction problem~\cite{jin2023spatio}, influenced
by both historical time-series patterns and spatial dependencies
across regions. 
However, our empirical analysis of DiDi’s operational data reveals that temporal dependencies exert a dominant
role, driven by human mobility patterns across daily, weekly, and seasonal cycles. Conversely, given that the spatial hierarchy of ride-hailing forecasting extends down to the county level, the geographic distance between counties and variations in their internal transportation conditions lead to a relatively limited impact on ride-hailing indicators of adjacent counties. Consequently, \textit{we formulate ride-hailing forecasting as a time series forecasting problem}.

As a specific application of time series forecasting, ride-hailing forecasting possesses a unique and pronounced characteristic: \textbf{\textit{Geo\mbox{-}\\spatial Heterogeneity}}.
As illustrated in the left panel of Figure \ref{fig:intro},
three regions with diverse urban contexts exhibit varying Call patterns that are correlated with regional factors encompassing functionality, economics, and demographics.
From our industrial practice, a region’s identity can be characterized by two fundamental aspects, as shown in the right panel of Figure~\ref{fig:intro}: \textbf{(1) Semantic Attribute View}. 
The static identity of a region is characterized by its semantic attributes, exemplified by the distribution of Points-of-Interest (POIs),
which encapsulate its functional characteristics and operational purpose.
\textbf{(2) Temporal Mobility Pattern View}. The dynamic essence of a region is captured through temporal mobility patterns, which delineate its operational rhythm across various temporal cycles, from
intra-day patterns (e.g., morning / evening peaks) to weekly trends
(e.g., weekday / weekend shifts). 
Effectively synergizing these static and dynamic dimensions is crucial for achieving accurate ride-hailing forecasting.

In recent years, the advent of Large Language Models (LLMs)~\cite{achiam2023gpt,liu2024deepseek,comanici2025gemini} has begun to shift the paradigm in time series forecasting~\cite{jin2024position}. Characterized by their profound generalization abilities~\cite{budnikov2025generalization} and world-scale knowledge bases~\cite{manvi2023geollm}, LLMs exhibit a distinct advantage in this domain~\cite{jin2024position}. 
Nevertheless, in specialized domains with a scarcity of open-source data, such as ride-hailing forecasting, the direct application of these models faces significant challenges due to their inherent deficiency in specialized domain knowledge.
Therefore, to enhance the adaptability of LLMs for ride-hailing forecasting tasks, we propose augmenting them with the following two types of priors: 
\textbf{(1) Geospatial Heterogeneity Prior}.
As previously analyzed, geospatial heterogeneity, reflecting intrinsic disparities in urban functionality, economic profiles, and demographic composition, underpins the diverse demand patterns across a region.
Consequently, explicitly introducing this factor is crucial for developing more precise and context-aware forecasting models.
\textbf{(2) External Event Prior}. Ride-hailing supply-demand dynamics are shaped not only by the platform’s own operational cycles but also by exogenous events including weather, holidays, and special
activities (e.g., concerts, sporting events)~\cite{wgnn,refine,stresnet}.
The sporadic and often unpredictable nature of such events makes them notoriously difficult to model using methods that rely solely on historical pattern mining. Therefore, incorporating an External Event Prior stays essential for enhancing the responsiveness to external events of ride-hailing forecasting systems.

In this paper, we propose a \underline{M}ulti-\underline{V}iew \underline{G}eospatial \underline{R}epresentation Learning (\model) framework for Ride-Hailing Forecasting.
The proposed framework comprises two sequential stages: a pretraining stage designed to address the challenge of geospatial heterogeneity by learning comprehensive geospatial representations, and a subsequent forecasting stage that leverages these representations and further incorporates external events and textual descriptions to enhance ride-hailing forecast performance.
In the pretraining stage, we incorporate two complementary data modalities — POIs and temporal mobility patterns — to capture geospatial heterogeneity across different regions from both semantic attribute and temporal mobility pattern views.
A dual cross-attention mechanism, coupled with an attentional pooling module, produces the final comprehensive geospatial representations.
In the subsequent ride-hailing forecasting stage, through the integration of multi-view geospatial representation, an elaborated prompt generation network effectively identifies underlying and shared regional properties, and subsequently learns to adaptively utilize these properties to generate informative prompt features that enhance predictive performance.
Additionally, contextual factors are captured by integrating three key external variables: rainfall, holidays, and special events.

In summary, our contributions lie in the following aspects:
\vspace{-0.3em}
\begin{itemize}
[leftmargin=*]
    \item \textit{Multi-View Geospatial Representation Learning}. To the best of our knowledge, this is the first attempt to conduct pre-trained modeling from both semantic attribute view and temporal mobility pattern view to enhance ride-hailing forecasting.
    
    \item \textit{Prompt-Empowered Ride-Hailing
Forecasting}.
We propose a prompt-empowered framework for fine-tuning LLMs on the task of ride-hailing forecasting, which simultaneously incorporates both external factors and textual descriptions.
    \item \textit{Extensive empirical studies}. We conduct extensive experiments
over DiDi’s real-world datasets. 
The results demonstrate that our model achieves state-of-the-art performance with an average improvement of 1.8\% for Call and 1.5\% for TSH prediction.
    \item \textit{Practical deployment}. Our proposed method has been successfully deployed on the DiDi platform. 
We demonstrate the system's user interface, geospatial embedding vector library, and an Intelligent Subsidy Allocation experiment to showcase our practicality.

\end{itemize}

\begin{figure*}[!htbp]
    \centering
\includegraphics[width=\linewidth]
    {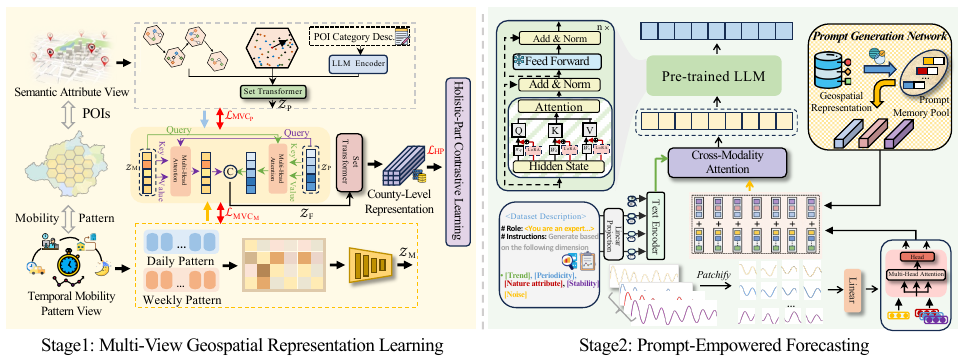}
    \vspace{-2em}
    \caption{The Overall Framework of~\model.}
    \label{fig:framework}
    \vspace{-1em}
\end{figure*}

\vspace{-0.5em}
\section{Preliminary}

\textbf{Problem Setting.}
In ride-hailing forecasting, we are given a historical ride-hailing series $X = \{x_{t-L+1:t}\} \in \mathbb{R}^{L \times N_c}$, where $L$ is the look-back window size, $N_c$ denotes the number of regions. Our goal is to learn a forecasting model $f( \cdot )$, which predicts the future $T$ time steps of the series, $\hat{X}=\{x_{t+1:t+T}\} \in \mathbb{R}^{T \times N_c}$, based on historical observations $X$.

Ride-hailing indicators can be classified into two categories: supply-side and demand-side.

\begin{itemize}
[leftmargin=*]
    \vspace{0.1em}
    \item \textbf{Call}: The total number of passenger-initiated ride requests within a specified time frame, \textit{acting as the core indicator of user \textbf{demand}}.
    \item \textbf{Total Supply Hour (TSH)}: The aggregate in-service duration of all drivers on the platform. \textit{It serves as a core indicator for measuring transport capacity  \textbf{supply}}.
\end{itemize}
We also incorporate three external variables identified through operational practice as most impactful on ride-hailing forecasting: \textbf{Rainfall}, \textbf{Holidays}, and \textbf{Special Events} (e.g., concerts and national examinations). More details can be found in Appendix~\ref{appendix:external}.

\section{Methodology}
In this section, we present details of our proposed~\model~in
Figure~\ref{fig:framework}, which consists of two main stages:
\begin{itemize}
[leftmargin=*]
    \item \textbf{Stage 1}: 
Our proposed Multi-View Geospatial Representation Learning framework comprises two complementary branches that model geospatial heterogeneity from semantic attribute view and temporal mobility pattern view, respectively.
The framework then facilitates cross-view adaptive interaction through a dual attention mechanism coupled with gated fusion operations.

\item \textbf{Stage 2}: 
In the forecasting phase, the time series data initially interacts with external variables and is then augmented with prompt-empowered geospatial representations from Stage 1. Subsequently, this enhanced representation undergoes cross-modal interaction with domain-specific text features. Finally, the resulting feature is fed into a pre-trained LLM, which is fine-tuned using Low-Rank Adaptation (LoRA) to generate predictions.

\end{itemize}
In industrial applications, the DiDi platform's ride-hailing forecasts currently operate fundamentally at the county level. 
However, China's county-level administrative units' size disparities (hundreds to thousands of square kilometers~\cite{countystatis}) hinder fine-grained geospatial heterogeneity modeling.
As a result, we commence our representation learning at a finer-grained grid level (hexagonal cells with 600m edge length) and subsequently aggregate incrementally upward to the county level.

\subsection{Multi-View Geospatial Representation Learning}

\noindent

\noindent
\textbf{Semantic Attribute View Modeling}.
The distribution of POIs reflects a region's functional characteristics and operational purpose~\cite{pgsimclr}.
To comprehensively model the semantic information behind POIs, we construct a multi-faceted representation by capturing their features from three diverse perspectives.
\ding{182} \textbf{Spatial Proximity}~\cite{place2vec}.
Leveraging a K-Nearest Neighbors (KNN) approach, we capture the distributional characteristics of POIs within their local spatial context. 
For each POI~~$p_{i}^{s}$, its $k$ nearest neighbors $p^{s}_{j} \in \mathcal{N}_k(p_i^s)$ are retrieved based on spatial distance. 
$c_i^s$ is the corresponding one-hot category vector associated with $p_{i}^{s}$,
which are then passed through an encoder $F_s(\cdot)$ to obtain their feature representations. 
The encoder consists of an embedding layer followed by a multi-layer perceptron (MLP).
We denote the final spatial proximity representation as $\mathcal{Z}{p_1} = F_s(c_i^s)$.
$p^{s}_{j}$ primarily contributes during training,
with details provided in Appendix~\ref{appendix:spatial_proximity}.
\ding{183} \textbf{Hierarchical Category Semantics}~\cite{huang2022estimating}.
To capture hierarchical semantic relationships between POI categories, we construct a POI graph where nodes are POIs and edges are spatially weighted. Random walks~\cite{randomwalk} are used to sample spatial co-occurring sequences. For each sequence, the first node is the target POI $p_i^h$, and the rest $p_j^h \in \mathcal{N}_k(p_i^h)$ form its context. Each POI is associated with its secondary category one-hot vector $c_i^h$, and encoded by $F_h(\cdot)$ to obtain feature representations.
We denote the final hierarchical  category semantics representation as $\mathcal{Z}_{p_2} = F_h(c_i^h)$.
$p_j^h$ primarily contributes
during training, with details provided in Appendix~\ref{appendix:Hierarchicalcategory}.
\ding{184} \textbf{Textual Semantics.} To extract the semantic characteristics of POI categories from a natural language perspective, we developed a specialized prompt that exploits the inherent knowledge embedded within LLMs. 
The prompt (Appendix~\ref{appendix:poi_prompt}) is specifically designed to generate comprehensive descriptions that precisely capture the distinct urban functional roles of individual POI categories. 
These generated POI descriptions are subsequently encoded by the LLM~\cite{gpt2} to obtain the feature representation $\mathcal{Z}_{P_3}$.
An attentional pooling mechanism~\cite{settransformer,yu2022coca} is then utilized to capture complex inter-feature dependencies and execute feature aggregation, ultimately yielding the final semantic attribute representation of grid-level region with $d$-dimension $\mathcal{Z}_{P} \in \mathbb{R}^{N_r \times d}$, where $N_r$ denotes the number of regions:
\begin{equation}
    \mathcal{Z}_{P} = \text{AttentionalPooling}(\text{Concat}(\mathcal{Z}_{P_1},\mathcal{Z}_{P_2},\mathcal{Z}_{P_3})).
\end{equation}

\noindent
\textbf{Temporal Mobility Pattern View Modeling}.
Given that daily cycles constitute the fundamental rhythm of urban activities and that weekdays exhibit distinct characteristics from weekends, these temporal factors significantly influence the fluctuation patterns in ride-hailing indicators. 
To capture these patterns, we compress the time series data of ride-hailing indicators within each region over a given period by computing averages along two temporal dimensions. \ding{182} Aggregating into 24-hour daily cycles to form a 24-dimensional vector; \ding{183} Aggregating into weekly cycles to form a 7-dimensional vector. 
We subsequently construct a joint hour-day matrix, with each element denoting the average indicator value for a specific hour within a particular day of the week. This methodology enables the extraction of fine-grained temporal mobility patterns, such as the potential differences between Monday morning peak hours and Sunday morning peak hours.
These temporal patterns are subsequently encoded into feature representations $\mathcal{Z}_{M} \in \mathbb{R}^{N_r \times d}$ through an encoder consist of MLP layers, followed by Multi-Head Attention~\cite{vaswani2017attention} to adaptively capture time-dependent patterns.

\noindent
\textbf{Cross View Adaptive Interaction \& Fusion}.
Subsequently, we fuse the semantic attribute features $\mathcal{Z}_{P}$ and temporal pattern features 
$\mathcal{Z}_{M}$ via a dual cross-attention mechanism~\cite{zhou2021informer,chen2021crossvit} to establish dynamic interactions between complementary domain features, yielding unified multi-view geospatial representations.
\begin{equation}
\mathcal{M}_{att} = \text{Softmax}\left(\frac{(W_Q\mathcal{Z}_{M}) (W_K\mathcal{Z}_{P})^T}{\sqrt{d}}\right) (W_V\mathcal{Z}_{P}),
\end{equation}
\begin{equation}
\mathcal{I}_{att} = \text{Softmax}\left(\frac{(W_Q\mathcal{Z}_{P}) (W_K\mathcal{Z}_{M})^T}{\sqrt{d}}\right) (W_V\mathcal{Z}_{M}),
\end{equation}
\vspace{0.2em}
\begin{equation}
    \mathcal{Z}_{F}= \text{Concat}(I_{att}, M_{att}),
\end{equation}
where $W_Q \in \mathbb{R}^{d \times d},W_K \in \mathbb{R}^{d \times d},W_V \in \mathbb{R}^{d \times d}$ are learnable matrices.
A multi-view consistency loss is utlized to enforce semantic consistency between $\mathcal{Z}_{F}$ and the single-view features $\mathcal{Z}_{P}$ and $\mathcal{Z}_{M}$.

\begin{equation}
\small
\mathcal{L}_{\text{MVC}_{\text{P}}} = -\log \frac{\exp(\operatorname{sim}(\mathcal{Z}_{Fi}, \mathcal{Z}_{P+})/\tau)}{\exp(\operatorname{sim}(\mathcal{Z}_{Fi}, \mathcal{Z}_{P+})+\sum_{j=0}^{N_L-1} \exp(\operatorname{sim}(\mathcal{Z}_{Fi}, \mathcal{Z}_{P-})/\tau)},
\end{equation}
where $\mathcal{Z}_{P+}$ denotes the $\mathcal{Z}_{P}$ representation that resides in the same region as $\mathcal{Z}_{Fi}$, while 
$\mathcal{Z}_{P-}$ denotes the representation located in different regions.
$\text{sim}(\cdot)$ denotes cosine similarity.
$N_L$ stands for the number of negative samples selected from the batch.
$\tau$ is the temperature parameter.
\begin{equation}
    \small
    \mathcal{L}_{\text{MVC}_{\text{M}}} = -\log \frac{\exp(\operatorname{sim}(\mathcal{Z}_{Fi}, \mathcal{Z}_{M+})/\tau)}{\exp(\operatorname{sim}(\mathcal{Z}_{Fi}, \mathcal{Z}_{M+})+\sum_{j=0}^{N_L-1} \exp(\operatorname{sim}(\mathcal{Z}_{Fi}, \mathcal{Z}_{M-})/\tau)},
\end{equation}
where $\mathcal{Z}_{M+}$ denotes the $\mathcal{Z}_{M}$ representation that resides in the same region as $\mathcal{Z}_{Fi}$, while 
$\mathcal{Z}_{M-}$ denotes the representation located in different regions.

After obtaining the grid-level representations $\mathcal{Z}_F$, we derive the county-level representations through the attentional pooling fusion mechanism:
\begin{equation}
    \mathcal{H} = \text{AttentionalPooling}(\mathcal{Z}_F)
\end{equation}
where $\mathcal{H} \in \mathbb{R}^{N_c \times d}$, $N_c$ denotes the number of county.
The Holistic-Part Loss is designed to align a county's feature representation with those of its constituent grid cells, while simultaneously distinguishing it from the representations of grid cells in other counties:
\begin{equation}
    \mathcal{L}_{HP} = -\log \frac{\exp(\operatorname{sim}(\mathcal{H}_{i}, \mathcal{Z}_{Fj})/\tau)}{\exp(\operatorname{sim}(\mathcal{H}_{i}, \mathcal{Z}_{Fj})+\sum_{k=0}^{N_L-1} \exp(\operatorname{sim}(\mathcal{H}_{i}, \mathcal{Z}_{Fk})/\tau)},
\end{equation}
where 
$\mathcal{Z}_{Fj}$ denotes the $\mathcal{Z}_{F}$ representation that resides in $\mathcal{H}_{i}$ area, while $\mathcal{Z}_{Fk}$ denotes the grid representation located in
different counties,
$N_L$ denotes the number of negative samples from the batch.

\subsection{Prompt-Empowered Forecasting}

In Stage 2, we integrate the comprehensive geospatial representation learned from Stage 1 into the forecasting process, incorporating county-level heterogeneity prior.

\subsubsection{Ride-Hailing Forecasting with Exogenous Factors}
In the context of ride-hailing practice, the supply-demand dynamics are governed by an interplay of endogenous factors, such as the platform's operational cycles, and exogenous variables like weather, holidays, and special events (e.g., concerts, sporting events). Accurately forecasting ride-hailing indicators thus necessitates a model capable of capturing the complex relationships between these two types of influences. To this end, we leverage the semantic understanding capabilities of LLMs~\cite{exollm} to interpret and encode the impact of these endogenous and exogenous factors on ride-hailing patterns.

As illustrated in the right panel of Figure~\ref{fig:framework}, for the ride-hailing indicator with input length $L$, ~$X \in \mathbb{R}^{N_c \times L}$, where $N_c$ denotes the number of counties, together with three exogenous variables — rainfall $X_r \in \mathbb{R}^{N_c \times L}$, holiday $X_h \in \mathbb{R}^{L}$, and special events $X_e \in \mathbb{R}^{N_c \times L}$ — we first split the time series into non-overlapping patches, and then performs self-attention interactions through a [EOS] token that aggregates global information.
\begin{equation}
    \hat{\textbf{x}} = \pi_N(
    \text{MLP}(\text{MSA}(\textbf{x} + p))).
\end{equation}
where $\text{MSA}(\cdot)$ denotes multi-head attention applied to time series, $p$
represents the position embedding, $\pi_N(\cdot)$ denotes the projection operation for selecting the last patch, $\text{\textbf{x}} \in \{X, X_r, X_h, X_e\}$, $\hat{\text{\textbf{x}}} \in \{\hat{X}, \hat{X_r}, \hat{X_h}, \hat{X_e}\} \in \mathbb{R}^D$, where $D$ denotes feature dimension.

After acquiring time series features $\hat{\textbf{x}}$,
we proceed to interact them with textual features and fuse the endogenous ride-hailing features with geospatial representation priors, then utilizing a pretrained LLM~\cite{jin2024position} to encode the combined representations. 
The textual descriptions of the time series data represent a critical component in exploiting the prior knowledge embedded within LLMs.
We utilize text generation prompts from~\cite{exollm} to provide comprehensive descriptions of both ride-hailing indicators and external variables from five perspectives: nature attribute, trend, periodicity, stability, and noise.
The generated textual descriptions are encoded using an LLM encoder~\cite{gpt2,exollm} to obtain textual features $\mathcal{T}$.

\subsubsection{Heterogeneity-Informed Prompt Learning}
As depicted in the right panel of Figure~\ref{fig:framework}, to integrate the learned geospatial representations into the ride-hailing forecasting model, we employ a Prompt Generation Network to maintain a globally shared memory pool, which consists of $M$ Key-Value pairs:
\begin{equation}
    \text{MP} = \{(k_0,v_0),(k_1,v_1),...,(k_{M-1},v_{M-1})\},
\end{equation} 
where $\text{MP} \in \mathbb{R}^{M \times d}$, $(k_i$,~$v_i)$ are all learnable parameters.
These are continuously optimized throughout the entire training process to store universal geospatial patterns.
The keys will ultimately learn to become various distinct prototypes, with each key representing a type of urban area that shares similar urban characteristics and intrinsic rhythms. The values store a set of mobility behavioral biases tailored to specific spatio-temporal prototypes.

Within this network, the learned geospatial representation functions as the query, which is used to match all Keys in the prompt memory pool, selecting the $k_p$ highest-scoring candidates, yielding a set of attention weights. The computed attention weights are then used to perform a weighted summation of all corresponding Values in the memory pool. Specifically, for the geospatial representation $\mathcal{H}_r$ corresponding to region $r$, we have:
\begin{equation}
    \mathcal{P}_r = \sum_{j=0}^{k_p-1} \alpha_j~v_j,~~~
\{\alpha_i\}_{i=0}^{k_p-1} = 
\mathop{\arg\max}_{l~\subseteq~[0,~M-1]}  \gamma\bigl(\mathcal{H}_r,\, k_l\bigr),
\end{equation}
where $\gamma(\mathcal{H},\, k)$ calculates cosine similarity.
Hereby, we establish a dynamic mapping from geospatial representations to model mobility behavioral heterogeneity, enabling the model to flexibly adapt to the ever-changing urban spatio-temporal scenarios, thus achieving exceptional generalization capability.

Finally, we concatenate the obtained prompt features ~$\mathcal{P} \in \mathbb{R}^{N_c \times d}$ and ride-hailing time series features to get $\mathcal{U} \in \mathbb{R}^{N_c \times d}$, and then interact them with text features. 
\begin{equation}
F = \text{Softmax}\left(\frac{(W_Q\mathcal{T}) (W_K\mathcal{U})^T}{\sqrt{d}}\right) (W_V\mathcal{U}),
\end{equation}
where $W_Q$,$W_K$,$W_V$ are learnable matrices.
Since the fusion of pre-trained geospatial representations with time series features differs from pure temporal feature distributions and semantic structures, unlike \cite{timellm,exollm}, we adopt LoRA~\cite{lora} to perform efficient fine-tuning in a parameter-efficient fine-tuning manner, thereby better leveraging the introduced spatiotemporal prior knowledge to enhance prediction performance.
LoRA facilitates the adaptation of Large Models (LMs) by injecting trainable, low-rank matrices into their Transformer layers to approximate weight updates. For a given pre-trained weight matrix $\textbf{W} \in \mathbb{R}^{d \times k}$, its update $\Delta W$ is represented by a low-rank factorization: $\textbf{W} + 
\Delta W = \textbf{W} + \textbf{W}_{\text{down}} \textbf{W}_{\text{up}}$. Here, $\textbf{W}_{\text{down}} \in \mathbb{R}^{d \times r}$ and $\textbf{W}_{\text{up}} \in \mathbb{R}^{r \times k}$ are the trainable low-rank matrices, with $r \ll \text{min}(d,k)$. 
This method is specifically applied to the query ($W_Q$), key ($W_K$) and value ($W_V$) projection matrices in the multi-head attention sub-layer, modifying the output projection h for any given input $x$.
\begin{equation}
    h = h + \varsigma\cdot xW_{\text{down}}W_{\text{up}},
\end{equation}
where $\varsigma \geq 1$  is a tunable scalar hyperparameter.

\vspace{-0.2em}
\section{Experiments}
In our experiments, we aim to address the following research questions (RQs):
\begin{itemize}
[leftmargin=*]
    \item \textbf{RQ1}: Can~\model~outperform prior  approaches under DiDi’s real-world oper-
ational datasets? $\Rightarrow$ \textbf{Sec. 4.2}.

    \item \textbf{RQ2}:  What are the individual contributions of the various components of~\model~to its overall effectiveness?$\Rightarrow$ \textbf{Sec. 4.3}.
    \item \textbf{RQ3}:  What does qualitative analysis reveal about the performance and interpretability of~\model? $\Rightarrow$ \textbf{Sec. 4.4}.
    \item \textbf{RQ4}:  How is the practical application of~\model~in real-world business? $\Rightarrow$ \textbf{Sec. 4.5}.
\end{itemize}

\begin{table*}[]
    \caption{Ride-hailing indicators prediction results. The best results are in bold and the second-best results are \underline{underlined}.}
    \vspace{-1em}
    \setlength{\tabcolsep}{0.8mm}{} 
    \begin{small}
    \scalebox{0.85}{
    \begin{tabular}{c|c|c|cc|cc|cc|cc|cc|cc|cc|cc|cc} 
    
    \toprule
    
    \multicolumn{3}{c|}{Methods} & \multicolumn{2}{c|}{\cellcolor{cyan!3}Ours} & \multicolumn{2}{c|}{\cellcolor{orange!3}ExoLLM} & \multicolumn{2}{c|}{\cellcolor{orange!3}iTransformer} & \multicolumn{2}{c|}{\cellcolor{orange!3}PatchTST} & \multicolumn{2}{c|}{\cellcolor{orange!3}DLinear}& \multicolumn{2}{c|}{\cellcolor{orange!3}CrossFormer} & \multicolumn{2}{c|}
    {\cellcolor{red!5}XGB} & \multicolumn{2}{c|}
    {\cellcolor{magenta!3}\cellcolor{red!5}Weekly
    Counterpart} & \multicolumn{2}{c}{\cellcolor{red!5}ARIMA} \\
     
    \midrule
    
    \multicolumn{3}{c|}{Metric} & $\text{WMAPE}$ & MAE & $\text{WMAPE}$ & MAE & $\text{WMAPE}$ & MAE & $\text{WMAPE}$ & MAE & $\text{WMAPE}$ & MAE & $\text{WMAPE}$ & MAE & $\text{WMAPE}$ & MAE & $\text{WMAPE}$ & MAE &$\text{WMAPE}$ & MAE
    \\
    
    \midrule
    \multirow{4}{*}{\rotatebox{90}{2025}}
    & \multirow{2}{*}{Call} & C1 & \textbf{0.114} & \textbf{48.158} & \underline{0.141} & \underline{59.377} &
    0.192& 74.621 &
    0.210 & 81.321 & 0.213 & 81.389 & 
    0.178 &69.042& 0.165 & 70.155 &
    0.293 & 124.262 &
     0.191&
81.085 \\
    
    & & C3 & \textbf{0.089} & \textbf{50.762} & \underline{0.105} & \underline{60.244} & 
   0.127& 67.520&
    0.139 & 73.903 & 
    0.139 &73.751& 
   0.122 &64.676& 
    0.127 & 78.937 &
    0.170 &97.263&
     0.125&
71.359 \\
     
    \cmidrule(lr){2-21} 
    & \multirow{2}{*}{TSH} & C1 & \textbf{0.027} & \textbf{14.419} & \underline{0.043} & \underline{22.664} &
    0.061&
29.447 &
    0.070 & 34.509 & 0.070&
33.966
 & 0.062&
29.976 & 0.057&
29.908 &
    0.075&
39.583 &
      0.059&
30.869 \\
    
    & & C3 & \textbf{0.039} & \textbf{15.414} & \underline{0.055}&
\underline{21.773}&
0.089&
32.592 & 
  0.092&33.804 &
   0.091&
33.434& 
   0.085&
31.390 & 
    0.089&
34.879&
    0.124&
48.844 &
     0.090&
35.239 \\
    
    \midrule
    \multirow{4}{*}{\rotatebox{90}{2024}}
    & \multirow{2}{*}{Call} & C1 & \textbf{0.104}& \textbf{42.762} & \underline{0.122} & \underline{50.083} &
    0.153& 55.924 &
     0.166 & 60.587 & 0.170&
60.680& 
   0.137 & 49.932 & 
    0.190 & 80.643 & 
     0.194 & 82.507 &
     0.193&
81.462 \\
    
    & & C3 & \textbf{0.078} & \textbf{47.662} & \underline{0.089} & \underline{54.230} &
    0.111& 61.956 &
    0.119 & 66.266 & 0.118&
66.069 & 
    0.106& 59.056 & 0.124 & 71.126 &
   0.142& 88.115 &
     0.135&
80.278\\
     
    \cmidrule(lr){2-21} 
    & \multirow{2}{*}{TSH} & C1 & \textbf{0.028} & \textbf{14.196} & \underline{0.029}&
\underline{14.403}&
0.052&
23.505&
     0.059&26.388 & 0.057&
25.593 & 
   0.055&
24.866& 
    0.055&
27.721 & 
     0.053&
 26.501 &
     0.057&
26.283 \\
    
    & & C3 & \textbf{0.032} & \textbf{13.122} & \underline{0.050}&
\underline{20.405}&
0.070&
26.384 &
    0.075&28.275 & 0.074&
27.820& 
    0.070&
26.220 & 0.077&
31.306 &
   0.083&
33.886 &
    0.075&
30.740 \\
    
    \midrule
    \multirow{4}{*}{\rotatebox{90}{2023}}
    & \multirow{2}{*}{Call} & C1 & \textbf{0.097} & \textbf{44.548} & \underline{0.116} & \underline{53.154} &
    0.157& 63.842&
    0.168 & 68.462 & 
    0.167& 68.333 & 
    0.151& 61.462 & 0.178 & 76.432 &
    0.217& 100.558 &
    0.183&
83.225 \\
    
    & & C3 & \textbf{0.088} & \textbf{41.611} & \underline{0.103} & \underline{48.673} &
    0.136& 58.143 &
    0.145 & 62.200 &
    0.147& 62.112 & 
    0.128& 54.916& 0.145 & 64.902 &
    0.175& 82.709 &
     0.158&
71.324 \\
     
    \cmidrule(lr){2-21}
    & \multirow{2}{*}{TSH} & C1 & \textbf{0.028} & \textbf{13.520} & \underline{0.049}&
\underline{23.847}&
0.076&
33.250&
    0.083&
36.357 & 
    0.082&
35.887 & 
    0.074&
32.418& 0.084 & 37.514 &
    0.090&
 44.740 &
     0.087 & 41.256 \\
    
    & & C3 & \textbf{0.040} & \textbf{13.583} & \underline{0.057}&
\underline{19.395}&
0.097&
29.342 &
   0.099&30.101 &
   0.098&
29.814 & 
    0.092&
27.984 & 0.102 & 33.070 &
   0.131&
44.168 &
     0.114 & 36.251 \\
    
    \midrule
    
    \multicolumn{3}{c|}{\cellcolor{gray!10} $1^{\text{st}}$ Count} & \multicolumn{2}{c|}{\cellcolor{gray!10} 24} & \multicolumn{2}{c|}{\cellcolor{gray!10} 0} & \multicolumn{2}{c|}{\cellcolor{gray!10} 0} & \multicolumn{2}{c|}{\cellcolor{gray!10} 0} & \multicolumn{2}{c|}{\cellcolor{gray!10} 0} & \multicolumn{2}{c|}{\cellcolor{gray!10} 0} & \multicolumn{2}{c|}{\cellcolor{gray!10} 0} & \multicolumn{2}{c|}{\cellcolor{gray!10} 0} & \multicolumn{2}{c}{\cellcolor{gray!10} 0}
    \\
    
    \bottomrule
    \end{tabular}
    }
    \label{tab:results_overall}
    \vspace{-1em}
    \end{small}
\end{table*}

\vspace{-0.2em}
\subsection{Experimental Setup}
\subsubsection{Datasets}
We study our problem on DiDi’s real-world operational datasets
encompassing two ride-hailing indicators: Call and TSH.
We focus on two core service
categories: C1 (Regular Express) and C3 (Economy Express).
The data is collected from 392 key counties of business interest
across China, with records taken at 30-minute intervals. To ensure broad temporal span coverage, we select data spanning from 2023, 2024 and 2025. The division of the dataset into train, val, and test sets is shown in Table~\ref{tab:dataset}.
The effective time window for evaluation is from 6:00 to 22:30 each
day.
We also incorporate external variables including rainfall, holidays, and special events, with temporal coverage consistent with the ride-hailing indicators in the dataset. Specifically, rainfall and special events are county-specific, while holidays are nationally uniform.

\begin{figure}[htbp]
    \centering
\includegraphics[width=\linewidth]
    {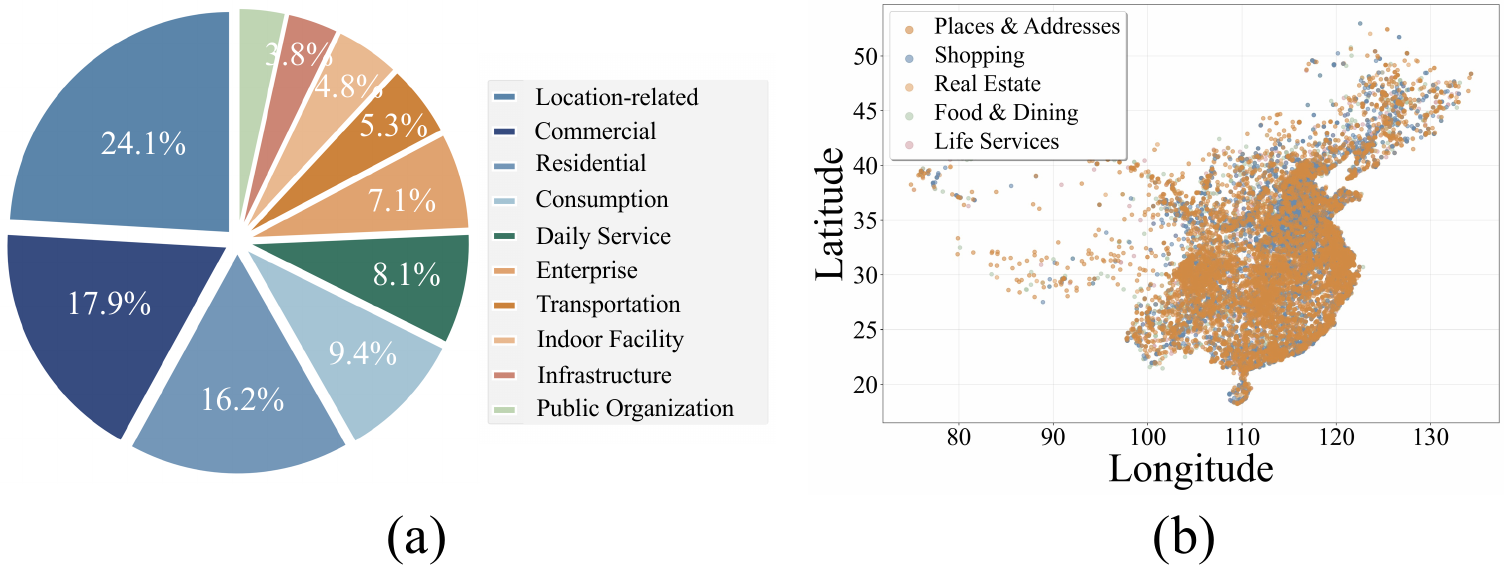}
    \vspace{-2em}
    \caption{The proportional distribution of top-10 and geographic distribution of top-5 POI primary categories.
    } 
    \label{fig:poi}
    \vspace{-1.5em}
\end{figure}

The dataset employed for pre-training consists of 93,441,589 POI entries with nationwide coverage across China. These entries are classified according to a two-tiered categorical system, which includes 18 primary and 218 secondary categories.
Figure~\ref{fig:poi} demonstrates the proportional distribution and geographic distribution of POI primary categories. 
More details can be found in Appendix~\ref{appendix:dataset}.

\textit{Ride-hailing forecasting faces an inherent data insufficiency problem}. Given that COVID-19's unprecedented impact~\cite{gao2020mapping} induced highly irregular consumption patterns during 2020-2022, data from this pandemic period proves unsuitable for operational forecasting applications. Additionally, earlier data from before 2019 represent outdated economic environments incompatible with present-day conditions. Consequently, the effective volume of applicable historical data remains severely constrained.

\subsubsection{Baselines}
We compare \model~with the following baselines on our dataset: deep learning based models like
PatchTST~\cite{patchtst}, DLinear~\cite{dlinear}, CrossFormer~\cite{zhang2023crossformer} and LLM-based
model ExoLLM~\cite{exollm}. In addition, we also compared the traditional
statistical learning method: XGB~\cite{chen2016xgboost}, ARIMA~\cite{shumway2017arima} and Weekly Counterpart.
Weekly Counterpart utilizes the value from the corresponding day of the preceding week, serving as a common baseline in industrial applications, as ride-hailing time series data characteristically exhibit strong daily and weekly seasonality.
Building upon the original models, we also introduce external variables as supplementary input features.

\vspace{-0.5em}
\begin{table}[htbp]
\centering
\caption{Dataset Division across training, validation, and test sets by time period.}
\label{tab:dataset}
\vspace{-1em}
\begin{adjustbox}{width=\columnwidth, center}
\begin{tabular}{cccccc}
\toprule
\textbf{Date} & \textbf{Train} & \textbf{Val} & \textbf{Test} & \textbf{Frequency} & \textbf{\# Counties} \\
\midrule
2023 & 01/01-05/15 & 05/16-05/31 & 06/01-06/30 & half-hour & 392 \\
2024 & 07/01-11/15 & 11/16-11/30 & 12/01-12/30 & half-hour & 392 \\
2025 & 01/01-05/15 & 05/16-05/31 & 06/01-06/30 & half-hour & 392 \\
\bottomrule
\end{tabular}
\end{adjustbox}
\end{table}
\vspace{-1em}

\subsubsection{Metrics and Implementation}
To assess the prediction performance, we adopt two commonly used evaluation metrics: Weighted
Mean Absolute Percentage Error (WMAPE) and Mean Absolute
Error (MAE).
The parameter initialization follows the setting from~\cite{exollm}. Adam~\cite{kingma2014adam} optimizer is chosen to minimize the training loss during parameter learning. 
All experiments are conducted on Tesla P40 and RTX A6000 GPUs. 
In our experiments, we set the batch size to 128, number of prompt pairs to 512 with $k_p$ set to 128. We set the look-back window $L$ to 336, corresponding to the past week, and the future timestep $M$ to 48. We utilize DeepSeek-R1~\cite{guo2025deepseekr1} in POI category description generation.

\subsection{RQ1: Overall Performance}
To evaluate the effectiveness of our~\model~framework, we conduct a comparison with existing state-of-the-art methods in Table~\ref{tab:results_overall}.
As can be seen, our model outperforms the other baselines across all years and in both categories.
The performance gain is 1.7\%,1.5\%,2.2\% for Call and 1.9\%,1.0\%,1.6\% for TSH in 2023,2024,2025, respectively.
Among the baseline methods, the LLM-based model (ExoLLM) demonstrates a clear advantage. Traditional deep learning methods (iTransformer, PatchTST, DLinear, CrossFormer) slightly outperform machine learning and statistical models (XGB, Weekly Counterpart, ARIMA). This progression indicates the evolutionary path and potential of model architectures for time series forecasting.

\subsection{RQ2: Ablation Studies}
\begin{figure}[htbp]
    \centering
\includegraphics[width=\linewidth]
    {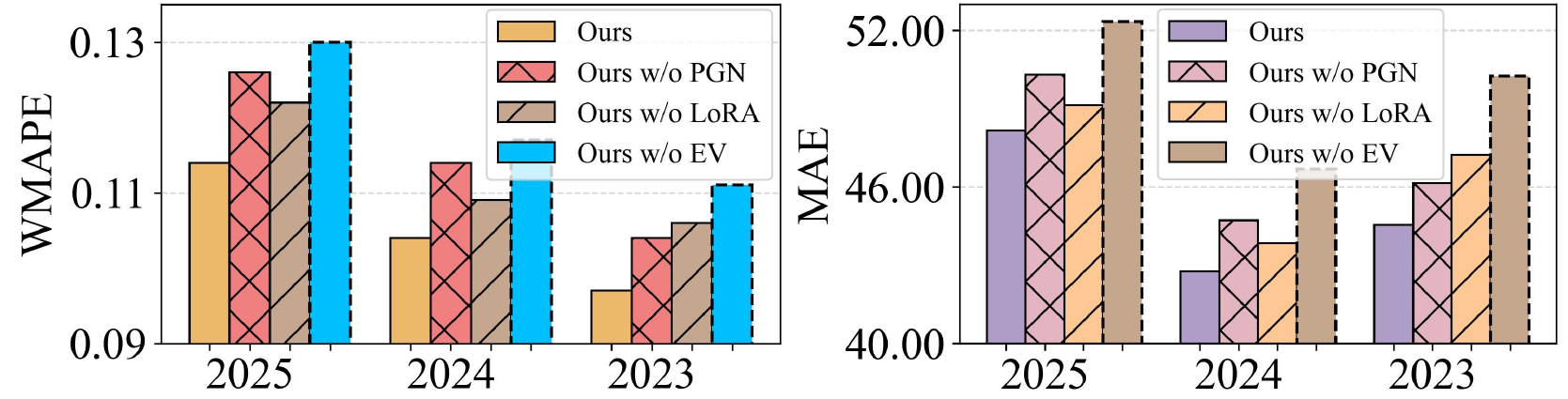}
    \vspace{-2em}
    \caption{Results of ablation studies on both WMAPE and MAE metrics. PGN stands for Prompt Generation Network, EV denotes External Variables.} 
    \label{fig:ablation}
\end{figure}

As shown in Figure~\ref{fig:ablation}, we conduct ablation studies to examine each component in~\model~on C1 category on our proposed dataset, including the Prompt Generation Network (PGN), LoRA adaptation, and External Variables (EV).
The ablation study reveals that all three components are integral to the performance of ~\model. 
Specifically,
without the Prompt Generation Network (PGN), the integration of geospatial representations becomes rigid and static, failing to capture dynamic spatial contexts.
Without LoRA, the LLM cannot adapt to the specific variations introduced by regional heterogeneity, thereby constraining its expressive capability for local patterns.
Without external variables, the model is unable to respond to the impact of external events in a timely manner.
Quantitatively, the inclusion of external variables consistently yields the most significant performance gain across all three years. Meanwhile, the relative contributions of the PGN and LoRA fluctuate, with their importance varying between different years.

\begin{figure}[htbp]
    \centering
\includegraphics[width=\linewidth]
    {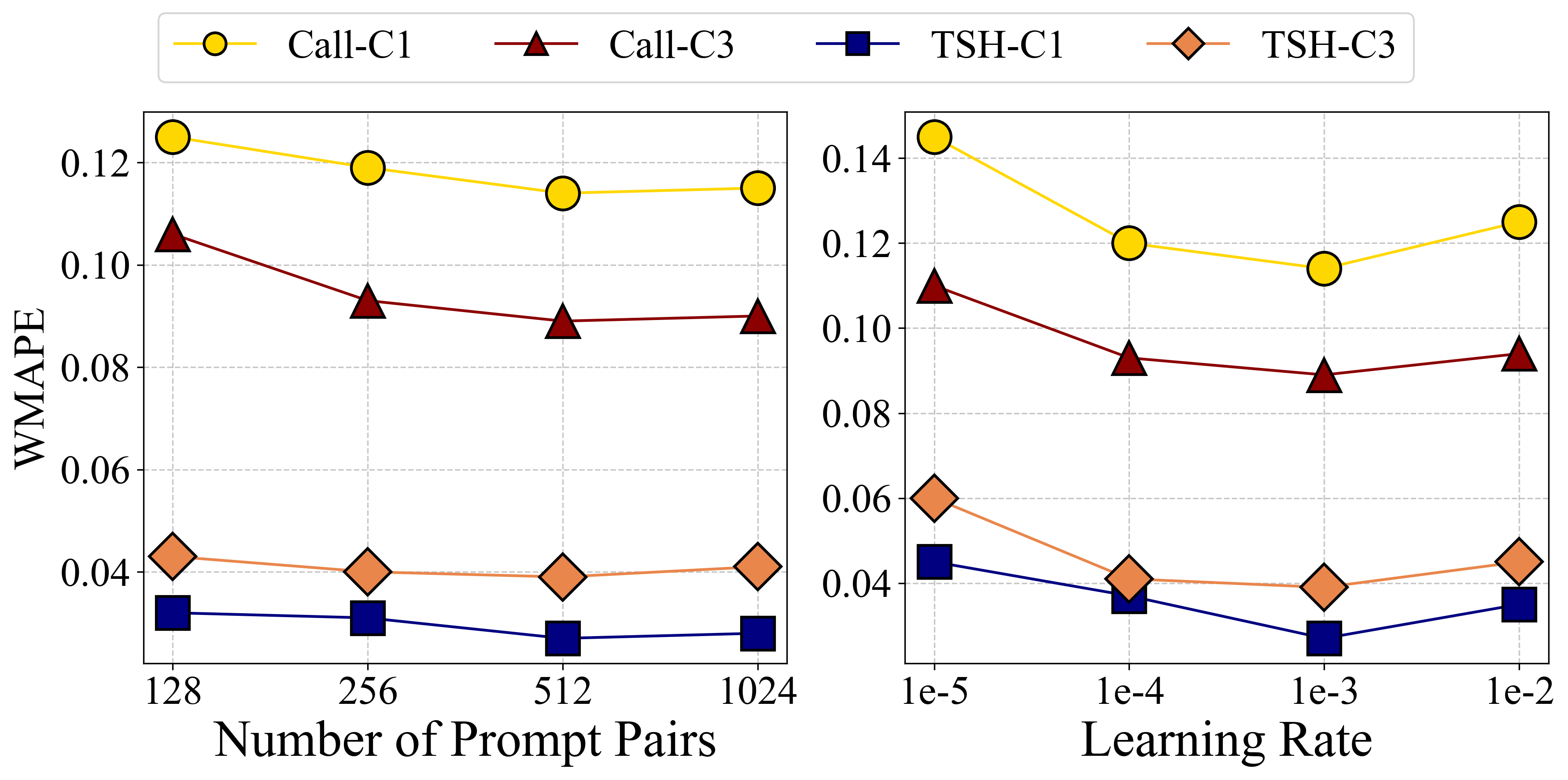}
    \vspace{-2em}
    \caption{Performance comparison for different hyperparameter setting on 2025.} 
    \label{fig:ablation2}
    \vspace{-1em}
\end{figure}

We further investigate the impact of key hyperparameters, including the number of prompt pairs in the Prompt Generation Network (PGN) and the learning rate, with the results presented in Figure~\ref{fig:ablation2}.
The experimental findings reveal a positive correlation between the number of prompt pairs and model performance, with performance gains plateau at 512 pairs.
Further increasing the number of prompt pairs to 1024 yields no additional performance improvement, suggesting that 512 pairs are sufficient to capture the essential patterns for our forecasting task.
The learning rate experiments exhibit a similar trend.
This analysis validates the effectiveness of our chosen hyperparameter settings.

\vspace{-1em}

\subsection{RQ3: Qualitative Analysis}
\subsubsection{Case Study for Predicted Results}
\begin{figure}[htbp]
    \centering
\includegraphics[width=\linewidth]
{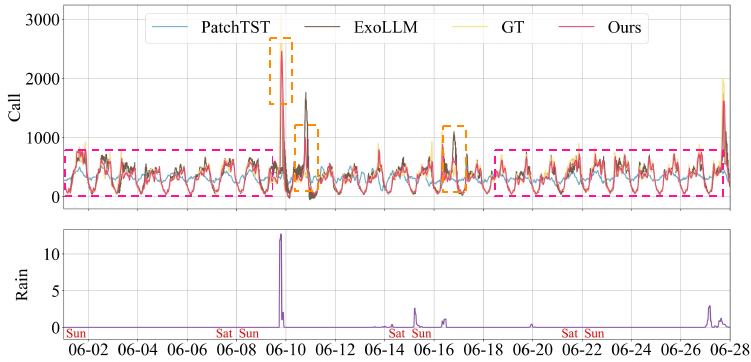}
    \vspace{-2em}
    \caption{Results of Region A210, Category C1 in 2025. GT denotes Ground Truth.}
    \label{fig:results}
    \vspace{-1em}
\end{figure}

To intuitively illustrate the performance of our proposed model, we present a visualization in Figure~\ref{fig:results} that compares its predicted call volumes against those of several baseline models.
To aid in understanding the fluctuations in the time series, the plot also includes recorded precipitation and indicates weekend periods.
As depicted, our method's predictions most closely align with the ground truth curve, accurately capturing both the underlying periodic trends (highlighted in \pink{pink}) and the abrupt spikes (highlighted in \orange{orange}). This result highlights the superior capability of our model to effectively integrate geospatial heterogeneity with external variables.

\vspace{-1em}

\subsubsection{Geospatial Representation Visualization}
\begin{figure}[H]
    \centering
\includegraphics[width=\linewidth]
    {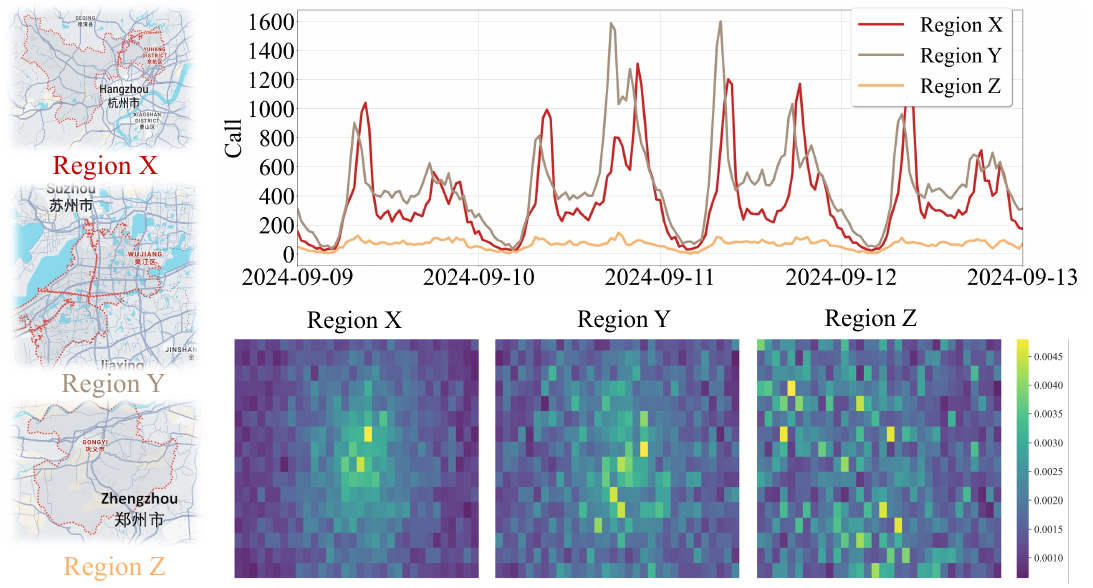}
    \vspace{-1em}
    \caption{Visualization of Prompt Memory Pool.}
    \label{fig:geo_repre}
    \vspace{-1em}
\end{figure}

In this section, 
to intuitively demonstrate the effectiveness of the geospatial representation learned in~\model,
we map the representation into two-dimensional space using the T-SNE algorithm~\cite{tsne} in Figure~\ref{fig:geo_repre}.
Subsequently, we employed the K-Means algorithm~\cite{kmeans} to partition the region representations into 10 clusters.
As we can see, the results reveal a high degree of clustering coherence, demonstrating that counties with similar attributes are effectively grouped together. 
For an intuitive and illustrative analysis, we highlight four representative clusters in detail. 
These clusters correspond precisely to key geo-economic archetypes within China: (1) Primarily Agricultural Counties,
(2) Underdeveloped and Ethnic Minority Inhabited Regions, (3) Remote, High-Altitude Cold Regions, and (4) Economically Strong Counties. 
It is noteworthy that the model successfully isolated the remote, high-altitude cold regions—areas empirically characterized by a significantly low volume of ride-hailing orders—as a unique and well-separated cluster.
This indicates that the learned geospatial representations successfully encapsulate the intrinsic attributes of the regions, functioning effectively as robust digital profiles for regional analysis.

\begin{figure}[H]
    \centering
\includegraphics[width=\linewidth]
    {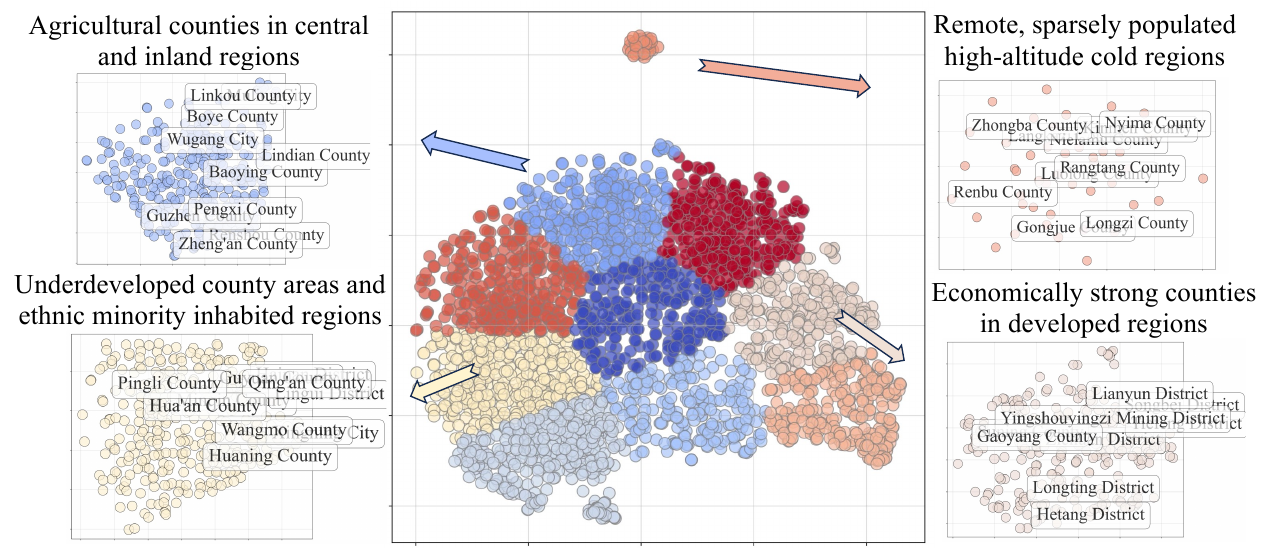}
    \caption{Geospatial Representation Visualization.}
    \label{fig:prompt_pool}
\end{figure}

\subsubsection{Visualization of Prompt Memory Pools}
We visualize the features learned by the prompt memory pool in Stage 2 to gain an intuitive understanding of its functionality in Figure~\ref{fig:prompt_pool}.
We selected three regions, Region X, Y, and Z, where X and Y are economically similar, developed regions located in China's coastal provinces, while Z exhibits a relatively lower level of economic development and is situated in the central inland area. 
The upper right panel displays the Call volume trends for these regions.
The bottom right figure illustrates the similarity scores between the geospatial representations of the corresponding regions and the 512 learnable Keys in the Prompt Generation Network, which we reshape into matrix form. As shown, the visualization matrices for Region X and Y exhibit similar distributions, which differ from the visualization matrix of Region Z.
This indicates that the learnable keys adaptively capture distinguishable socioeconomic characteristics.

\subsection{RQ4: Practicality}

\subsubsection{Deployment}
Our proposed \model~has been deployed in DiDi's operational environments, supporting multiple business scenarios such as demand-supply forecasting, resource allocation, smart subsidies, and dynamic pricing. To illustrate its practical use, we present a dashboard interface demonstration in Figure~\ref{fig:embedding_retrieval} in Appendix. The dashboard helps analysts and engineers visualize, monitor, and interactively analyze spatio-temporal mobility patterns. 
The data presented in the figure has been anonymized to protect sensitive information.

\subsubsection{Geospatial Embedding Vector Library}
\begin{figure}[H]
    \centering
\includegraphics[width=\linewidth]
    {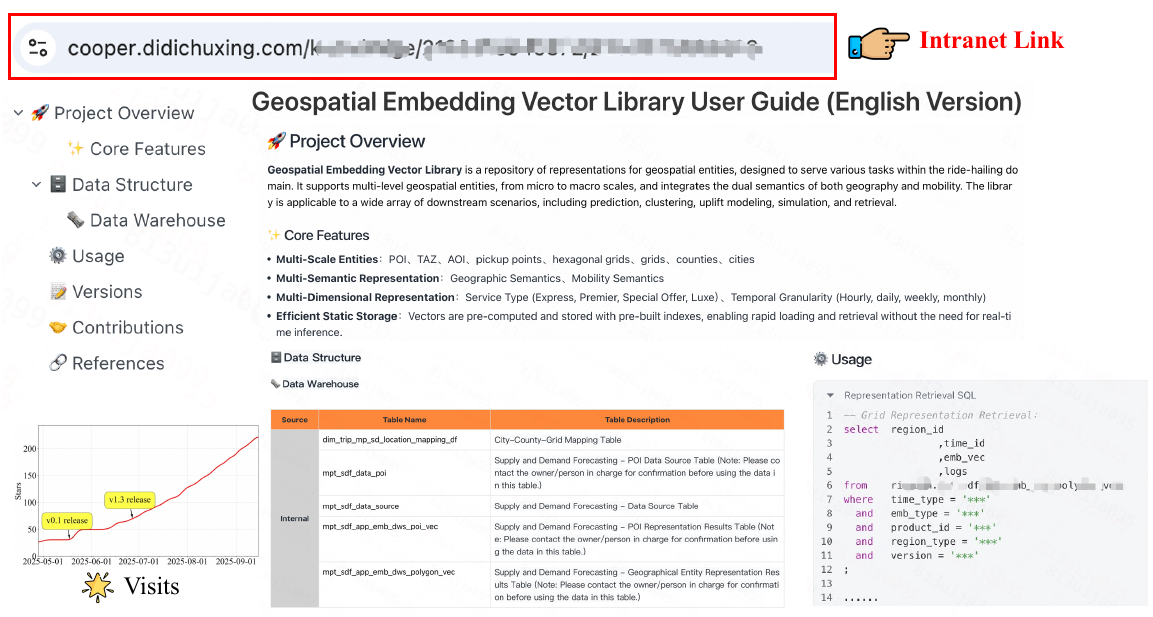}
    \vspace{-2em}
    \caption{Demonstration of internal document for Geospatial Embedding Vector Library.}
    \label{fig:geolibrary}
    \vspace{-1.5em}
\end{figure}

The geospatial representation we developed has been operationalized within DiDi as a production-ready geospatial embedding vector library. To date, it has undergone development across two generations, culminating in seven released versions.
Figure~\ref{fig:geolibrary} illustrates our internal web portal for the library, which provides key information such as the project overview, data structure, usage guidelines, and version history. 
The line chart in the bottom-left corner plots the growth in the number of visits that our embedding vector library has received since its release in April of this year.

\vspace{-1em}

\subsubsection{Online A/B Test}
Our proposed~\model~has been successfully deployed in live production environments to generate ride-hailing forecasts, thereby supporting downstream operational decision making processes.
We follow~\cite{hu2024enhancing,peng2024large,zou2024knowledge} to conduct online A/B testing to validate the effectiveness of~\model.
We report daily forecasting results conducted in August 2025.~\model~achieves average WMAPE improvement of 3.6\% and 2.3\% in Call performance for C1 and C3 categories respectively; 2.7\% and 2.8\% in TSH performance for C1 and C3 category respectively.

To further validate the practical utility of our model, we report its performance in a significant downstream task: \textit{Intelligent Subsidy Allocation}. 
Ride-hailing platforms commonly offer user subsidies to improve retention and operational efficiency. This task allocates subsidies based on demand forecasts to enhance overall conversion performance.
More details are provided in Appendix~\ref{appendix:subsidy}.

\section{Related Work}
\subsection{Urban Region Representation Learning.}
Learning representations of urban regions~\cite{hao2025unlocking,zou2025deep} targets the creation of highly transferable region embeddings via the incorporation of regions alongside their attributes.
Existing literature has leveraged various data modalities to capture the characteristics of regions, including visual imagery~\cite{hao2025urbanvlp,zou2024learning,satcle}, POI~\cite{xi2022beyond}, human mobility~\cite{zhou2023heterogeneous}, and knowledge graphs~\cite{liu2023knowledge}.
Pre-trained region features can be further adapted to various downstream tasks through fine-tuning, such as economic~\cite{hao2025urbanvlp,uspm,xiao2024refound}, environmental~\cite{zhang2025urban,chen2024terra}, and demographic applications~\cite{remvc}.
In this work, rooted in ride-hailing forecasting contexts, we combine POI data and temporal mobility patterns to model regional representations, infusing spatiotemporal heterogeneity into downstream applications.

\subsection{Ride-Hailing Forecasting}
In recent years, Ride-Hailing Forecasting~\cite{dmvstnet,stmgcn,ccrnn,pian2022spatialtemporaldynamicgraphattention,JIN202279,adformer,wgnn,zhang2021mlrnn,cao2021bert,li2024optimization,huang2022gan,lin2023deep,jin2020urban} has received widespread attention due to its alignment with practical industry demands.
~\cite{dmvstnet} integrates CNNs, LSTMs, and graph embeddings to capture spatial, temporal, and semantic regional dynamics. Regarding multi-relational learning, ~\citet{stmgcn} utilizes multi-graph convolution to represent inter-regional spatial, functional, and connectivity dependencies. For adaptive modeling, CCRNN~\cite{ccrnn} enables dynamic learning of location-specific adjacency matrices. To address noise mitigation, ADFormer~\cite{adformer} employs differential attention mechanisms at the architectural level for spatial correlation refinement.
In this work,
we formulate ride-hailing supply-demand forecasting as a time series forecasting problem, with detailed discussion in Section~\ref{text:intro}.

\subsection{Prompt Learning}
Prompt Learning~\cite{li2024flashst,yuan2024unist,dong2024heterogeneity,cao2024tempo,li2024opencity,yuan2024urbandit} suggests a methodology for efficiently adapting large-scale pre-trained models to downstream tasks by introducing a small number of trainable prompt parameters, and its influence has rapidly expanded from Natural Language Processing (NLP)~\cite{lester-etal-2021-power} to Computer Vision (CV)~\cite{sohn2023visual,zhou2022learning} and even multi-modal learning~\cite{khattak2023maple}. Furthermore, researchers have explored dynamic mechanisms that go beyond static prompts, such as leveraging a shared prompt pool to facilitate continual learning without rehearsal buffer~\cite{l2p}.
HimNet~\cite{dong2024heterogeneity} extracts spatio-temporal heterogeneity-informed prompt features through a query-pool paradigm.
In this work, we deeply integrates dynamic prompts with the problem of spatio-temporal heterogeneity by proposing a novel, retrieval-based adaptive prompt framework. 

\section{Conclusion and Future Work}
Accurate ride-hailing forecasting plays a pivotal role in the advancement of operational efficiency, the improvement of user experience, and the optimization of traffic management.
To tackle the geospatial heterogeneity challenge inherent in ride-hailing demand forecasting, this work proposes learning comprehensive geospatial representations through two complementary viewpoints: semantic attributes and temporal mobility patterns.
In subsequent ride-hailing forecasting, we integrate the geospatial representation into the time series forecasting through a dynamic prompt generation network, and combine it with external variables and domain-specific textual descriptions to enhance the prediction of ride-hailing indicators.
Experiments on DiDi business data demonstrate the effectiveness of our method. 
Future directions include exploring the application of other types of data such as origin-destination (OD) in ride-hailing forecasting, as well as applying novel backbones such as TabPFN~\cite{hollmann2025tabpfn}.

\begin{acks}
This work is supported by CCF-DiDi GAIA Collaborative Research Funds for Young Scholars, the National Natural Science Foundation of China (No.62402414), the Guangdong Basic and Applied Basic Research Foundation (No. 2025A1515011994), the Guangzhou Municipal Science and Technology Project (No. 2023A03J0011), the Guangzhou Industrial Information and Intelligent Key Laboratory Project (No. 2024A03J0628), and the Guangdong Provincial Key Lab of Integrated Communication, Sensing and Computation for Ubiquitous Internet of Things (No. 2023B1212010007).
\end{acks}

\balance

\clearpage
\appendix

\twocolumn[{
	\renewcommand\twocolumn[1][]{#1}
	\begin{center}
 \textbf{\fontsize{15}{48}\selectfont Appendix}
  \end{center}
        \vspace{0.5cm}
}]

\section{POI Prompt}
\label{appendix:poi_prompt}
We provide a detailed POI Category Description generation prompt below, which comprehensively explores POI semantic features from five perspectives: Core Function, Hierarchy, Target Demographics, Temporal Pattern, and Spatial Context.

\vspace{-1em}
\begin{figure}[htbp]
    \centering
\includegraphics[width=\linewidth]
    {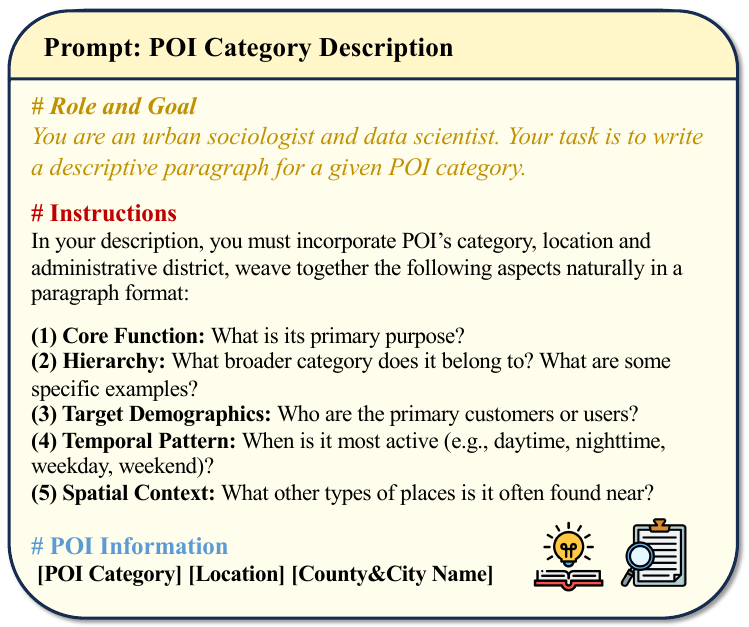}
    \vspace{-2em}
    \caption{Detailed prompt template for POI category description generation.}
    \label{appendix:poi_text}
    \vspace{-1.5em}
\end{figure}

\section{More Details about Dataset}
\label{appendix:dataset}

\textit{More information about POI secondary categories}.
For the POI data utilized in Stage 1: Multi-View Geospatial Representation Learning of our model, beyond the content introduced in the main text, we also introduce information about POI secondary categories here.
POI secondary categories provide a finer-grained functional breakdown of primary POI categories.
The top-10 POI secondary categories include:
Door Number Information, Building Numbers, Administrative Places, Companies \& Enterprises, Passage Facilities, Snacks \& Fast Food, Furniture \& Building Materials, Beauty \& Hair Salons, Government Agencies, Chinese Restaurants.

\textit{Special events data types}.
Special events data encompasses 12 distinct types:
\textit{International Women's Day, Teacher Qualification Examination, Public Institution Recruitment Examination, Provincial Civil Service Examination, Self-taught Higher Education Examinations, Chinese Valentine's Day, Art College Entrance Examination, Concert, Sports Event, Marathon, Beer Festival and Large-scale Exhibition}.
Some non-statutory holidays are also classified as special events.

\section{More Details about POI Representation}

\subsection{Spatial Proximity}
\label{appendix:spatial_proximity}
Here we introduce the training process of POI representation learning from spatial proximity.
For each POI~~$p_{i}^{s}$, and its $k$ nearest neighbors $p^{s}_{j} \in \mathcal{N}_k(p_i^s)$ are retrieved based on spatial distance. 
$c^s$ is the corresponding one-hot category vectors associated with $p^{s}$,
\begin{equation}
\mathbf{z}_{p_{i}^{s}} = F_s(c_i^s), \quad \mathbf{z}_{p_{j}^{s}} = F_s(c_j^s).
\end{equation}
The objective is to maximize the similarity between the central POI and its neighbors:
\begin{equation}
\mathcal{L}_{\text{SP}} =  -\sum_{p_{j}^s ~\in~ \mathcal{N}_k(p_i^s)}(\log \frac{\exp\left(\mathbf{z}_{p_{j}^{s}}^\top  ~\mathbf{z}_{p_{i}^{s}}\right)}{\sum_{l=0}^{n_c-1} \exp\left(\mathbf{e}_{l}^\top ~  \mathbf{z}_{p_{i}^{s}}\right)}),
\end{equation}
where 
$n_c$ is the total number of POI categories, $e_l$ denotes the feature representation of the $l$-th POI category after encoding through $F_s$.

\subsection{Hierarchical Category Semantics}
\label{appendix:Hierarchicalcategory}

Here we introduce the training process of POI representation learning from hierarchical category semantics.
To capture hierarchical semantic relationships between POI categories, we construct a POI graph where nodes are POIs and edges are spatially weighted. Random walks~\cite{randomwalk} are used to sample spatial co-occurring sequences. For each sequence, the first node is the target POI $p_i^h$, and the rest $p_j^h \in \mathcal{N}_k(p_i^h)$ form its context. Each POI is associated with its secondary category one-hot vector $c_i^h$, and encoded by $F_h(\cdot)$ to obtain feature representations:
\begin{equation}
\mathbf{z}_{p_{i}^{h}} = F_h(c_i^h), \quad \mathbf{z}_{p_{j}^{h}} = F_h(c_j^h).
\end{equation}
The hierarchical semantic representations are optimized via the following joint objective:
\begin{equation}
\begin{aligned}
\mathcal{L}_{\text{HCS}} &= -\sum_{p_{j}^{h} \in \mathcal{N}_k(p_i^h)}
\log \frac{\exp(\mathbf{z}_{p_{j}^{h}}^\top \mathbf{z}_{p_{i}^{h}})}
{\sum_{l=0}^{n_c-1} \exp(\mathbf{e}_l^\top \mathbf{z}_{p_{i}^{h}})} \\
&\quad + \lambda \sum_{i,l} w_{il} \left\| \mathbf{z}_{p_{i}^{h}} - \mathbf{e}_{l} \right\|_2^2,
\end{aligned}
\end{equation}
while the first term models spatial co-occurrence between categories using a skip-gram objective, 
the second regularization term enforces embedding smoothness for POIs within the same primary category, $n_c$ denotes the number of POI secondary categories, $w_{il}=1$ if $p_i^h$ and category $l$ belong to the same primary category group. $\lambda$ controls the trade-off between the two objectives.

\section{More Details about External Variables}
\label{appendix:external}
\begin{itemize}
[leftmargin=*]
    \item \textbf{Rainfall}. We incorporate rainfall data, measured as precipitation in millimeters (mm), as the sole weather variable due to its high impact. The forecasted rainfall over a prediction horizon of length $S$ is represented by the vector $\text{P} \in \mathbb{R}^{S×1}$
    where $\text{P} = (p_{t+1}, p_{t+2},...,p_{t+s})$
. Data is sourced from the China Weather Network (\url{https://www.weather.com.cn}).
\vspace{0.2em}
    \item \textbf{Holiday}.
    To capture the impact of public holidays on travel patterns, we incorporate holiday data as a categorical feature. We consider $N_h$ distinct types of public holidays known to significantly influence travel demand. This information is encoded as a vector for the forecast period:
\vspace{-0.2em}
\begin{equation*}
\mathbf{H} = (h_{t+1}, h_{t+2}, \dots, h_{t+S}) \in \mathbb{Z}^{S \times 1}, h_i \in [0, N_h],
\end{equation*}
where $S$ is the prediction horizon, $N_h$ correspond to the number of holiday types. 
The data is based on official holiday schedules announced by the Chinese government.
\vspace{0.2em}
    \item \textbf{Special Events}. Special events meta data accounts for non-periodic external events that can significantly impact travel demand. Examples include major concerts, sporting events, and national examinations.
\vspace{-0.2em}
\begin{equation*}
\mathbf{E} = (e_{t+1}, e_{t+2}, \dots, e_{t+S}) \in \mathbb{Z}^{S \times 1}, e_i \in [0, N_e],
\end{equation*}
where $N_e$ represents the number of special events.
Data is sourced from Damai (\url{https://www.damai.cn/}).
\end{itemize}

\section{More Details for Deployment}

\subsection{Deployment System Demonstration}

As introduced in the main text, the interface allows flexible spatial and temporal exploration. Users can enter a grid ID in the top-left map to query the most similar grids from the embedding database. It also supports analysis of POI functions within the grid, helping to understand regional functionality and potential travel demand. 
For prediction, users can select any province-level, city-level, or county-level region to monitor future CALL, TSH and other relevant indicators such as ASP (Average Selling Price). This helps guide resource allocation, pricing strategies, and operational planning.
\begin{figure}[h]
    \centering
\includegraphics[width=\linewidth]
    {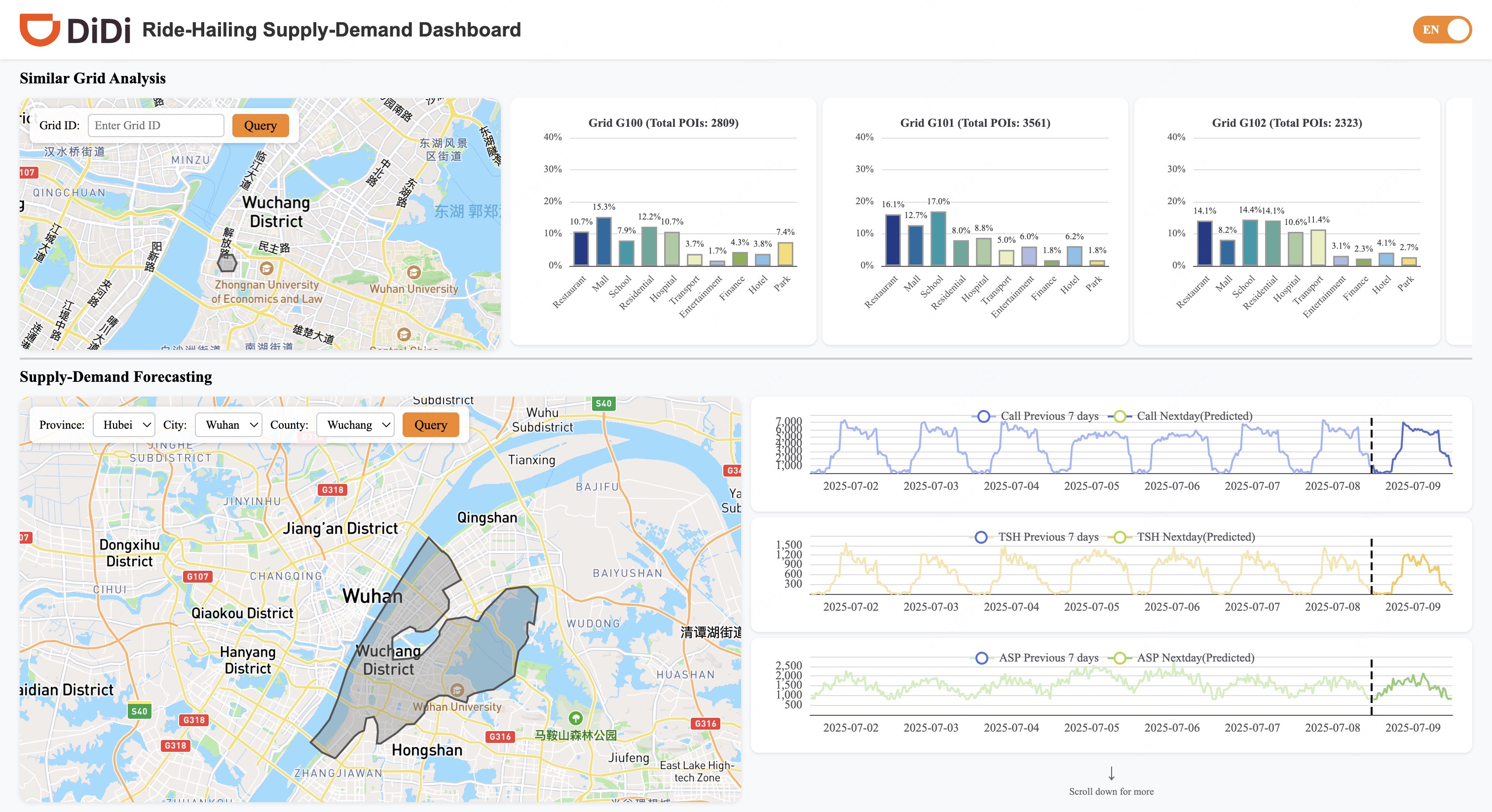}
    \vspace{-2em}
    \caption{Deployment system demonstration of our \model~framework.}
\label{fig:embedding_retrieval}
\end{figure}

\subsection{Intelligent Subsidy Allocation Experiment}
\label{appendix:subsidy}

To evaluate the transferability of the geospatial representations learned by \model~across multiple tasks, we additionally introduce an \textbf{Intelligent Subsidy Allocation Experiment}~\cite{yang2025decision}. 
To enhance operational efficiency and economic benefits, ride-hailing platforms typically provide subsidies to users in order to improve user retention.
Intelligent Subsidy Allocation aims to effectively allocate subsidies and improve overall operational efficiency and conversion performance. 
This task is formulated as a multi-treatment causal inference problem (\ie, multiple subsidy strategies). The objective is to estimate the heterogeneous impact of different subsidy strategies on user conversion behavior. We adopt a neural network architecture with multiple output heads, which allows the model to predict potential outcomes under all treatment conditions within a unified framework.

Let the strategy set be $\mathcal{T} = \{ST_0, ST_1, \dots, ST_{N_{st}-1}\}$, where $N_{st}$ denotes the number of strategies such as 10\% discount, fixed-amount reduction. $ST_0$ denotes no strategy. For input feature $\mathbf{x_{st}}$ of order and user interaction data, the model outputs the predicted conversion probability (the probability of a user taking a ride, conditional on receiving a subsidy) under each treatment: $\hat{Y}(ST_i\mid\mathbf{x_{st}})$. Each prediction is generated by a separate output head corresponding to one treatment.
During training, the model learns from historical samples $\Omega = (\mathbf{x_{st}}_i, ST_i, \hat{Y}_i)$, where $ST_i$ is the observed treatment and $\hat{Y}_i \in \{0, 1\}$ indicates whether the user converted. 
The cross-entropy loss function is used to optimize prediction accuracy under each treatment.
During inference, the model computes the uplift effect of any treatment $ST_i$ relative to the strategy $ST = 0$ as follows:
\begin{equation}
    U(ST_i \mid \mathbf{x_{st}}) = \hat{Y}(ST_i \mid \mathbf{x_{st}}) - \hat{Y}(ST_0 \mid \mathbf{x_{st}}).
\end{equation}
In business practice, user responses depend not only on order features but also on the spatial context, regional supply-demand conditions, and functional attributes. 
Therefore, we incorporate our proposed geospatial representations as features to capture these influences.
We augment the feature vector for each order by concatenating it with the pre-trained geospatial representation of its originating region, yielding the final model input.
The experiment results are demonstrated in Table~\ref{tab:subsidy_exp}.
The online baseline model employs a neural network with multiple output heads.
Based on this, the enhanced model $\text{Exp}_1$ utilized our city-level region representations, while $\text{Exp}_2$ utilized our county-level region representations. 
QINI coefficient~\cite{belbahri2021qini} and WMAPE are utilized to measure how well the model identifies high-response users.
Results demonstrate that the enhanced model almost significantly outperforms baselines on both QINI coefficient and WMAPE metrics. 
Notably, optimal performance varies by strategy, with $\text{Exp}_1$ excelling in some scenarios and  $\text{Exp}_2$ in others.
These results confirm the effectiveness and transferability of our learned geospatial representations.

\begin{table}[htbp]
\centering
\footnotesize  
\begin{tabularx}{0.45\textwidth}{c|*{3}{>{\centering\arraybackslash}X}|*{3}{>{\centering\arraybackslash}X}}
\toprule
& \multicolumn{6}{c}{\normalsize\textbf{2025 / 05}} \\
\cmidrule(lr){2-7}
\multirow{2}{*}{\textbf{Treatment}} & \multicolumn{3}{c|}{\textbf{QINI} $\downarrow$} & \multicolumn{3}{c}{\textbf{WMAPE} $\downarrow$} \\
\cmidrule(lr){2-4} \cmidrule(lr){5-7}
& Exp\_1 & Exp\_2 & Online & Exp\_1 & Exp\_2 & Online \\  
\midrule
85\%-x & 0.239 & \textbf{0.220} & 0.239 & \textbf{0.167} & 0.192 & 0.248 \\
80\%-x & 0.233 & 0.232 & \textbf{0.229} & \textbf{0.110} & 0.152 & 0.307 \\
75\%-x & 0.241 & \textbf{0.238} & 0.239 & \textbf{0.063} & 0.126 & 0.126 \\
70\%-x & \textbf{0.243} & 0.252 & 0.249 & 0.115 & \textbf{0.111} & 0.142 \\
60\%-x & 0.247 & \textbf{0.245} & \textbf{0.245} & \textbf{0.038} & 0.086 & 0.121 \\
\bottomrule
\end{tabularx}
\vspace{0.5em}
\caption{Geospatial representation enhanced intelligent subsidy experiment results for May 2025. Treatment name's first number represents the discount percentage, x stands for a direct reduction discount of x yuan. We set x to 5 here.}
\label{tab:subsidy_exp}
\end{table}
\vspace{-0.5em}

\end{document}